\documentclass[10pt,twocolumn,letterpaper]{article} 

\usepackage{cvpr}
\usepackage{times}
\usepackage{epsfig}
\usepackage{graphicx}
\usepackage{amsmath}
\usepackage{amssymb}
\usepackage{caption}

\usepackage{bm}

\usepackage[normalem]{ulem}

\renewcommand{\vec}[1]{\bm{#1}}
\newcommand{\myb}{\vec{\beta}}
\newcommand{\myt}{\vec{\theta}}
\newcommand{\myg}{\vec\gamma}
\newcommand{\vx}{\textbf{x}}

\usepackage[breaklinks=true,bookmarks=false]{hyperref}

\cvprfinalcopy 

\def\cvprPaperID{2825} 

\ifcvprfinal\fi
\begin{document}

\title{3D Menagerie: Modeling the 3D Shape and Pose of Animals}
\author{Silvia Zuffi$^1$\hspace{0.05\linewidth} Angjoo Kanazawa$^2$
\hspace{0.05\linewidth} David Jacobs$^2$ \hspace{0.05\linewidth}Michael J. Black$^3$\\
$^1$IMATI-CNR, Milan, Italy, 
$^2$University of Maryland, College Park, MD\\
$^3$Max Planck Institute for Intelligent Systems, T\"{u}bingen, Germany\\
{\tt\small silvia@mi.imati.cnr.it},
{\tt\small \{kanazawa, djacobs\}@umiacs.umd.edu},
{\tt\small black@tuebingen.mpg.de}
}

\twocolumn[{%
\renewcommand\twocolumn[1][]{#1}%
\maketitle
\begin{center}
  \newcommand{\teaserwidth}{0.98\textwidth}
 \vspace{-0.3in}
    \centerline{
    \includegraphics[width=\teaserwidth,clip]{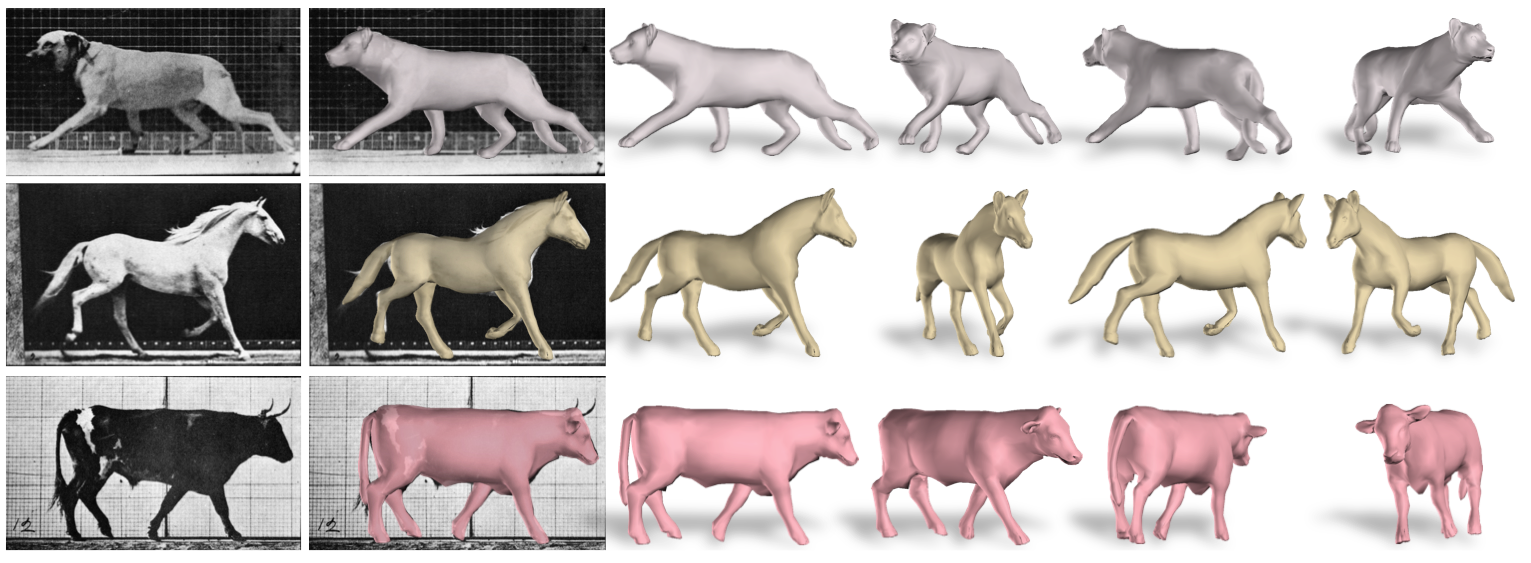}
    }
    \captionof{figure}{{\bf Animals from images}. 
We learn an articulated, 3D, statistical shape model of animals using
very little training data.
We fit the shape and pose of the model to 2D image cues showing how 
it generalizes to previously unseen shapes.
}
\label{fig:teaser}
\end{center}%
}]

\maketitle

\begin{abstract}
  \vspace{-.2em}

There has been significant work on learning realistic, articulated, 3D models of the human body. In contrast, there are few such models of animals, despite many applications. The main challenge is that animals are much less cooperative than humans. The best human body models are learned from thousands of 3D scans of people in specific poses, which is infeasible with live animals. 
Consequently, we learn our model from a small set of 
3D scans of {\em toy} figurines in arbitrary poses. We employ a novel part-based shape model to compute an initial registration to the scans. We then normalize their pose, learn a statistical shape model, and refine the registrations and the model together. In this way, we accurately align animal scans from different quadruped families with very different shapes and poses.
With the registration to a common template we learn a shape space representing animals including lions, cats, dogs, horses, cows and hippos. Animal shapes can be sampled from the model, posed, animated, and fit to data. 
We demonstrate generalization by fitting it to images of real animals including species not seen in training.
\end{abstract}

\section{Introduction}

The detection, tracking, and analysis of animals has many applications in biology, neuroscience, ecology, farming, and entertainment.
 Despite the wide applicability, the computer vision community has focused more heavily on modeling humans, estimating human pose, and analyzing human behavior.
Can we take the best practices learned from the analysis of humans and apply these directly to animals?
To address this, we take an approach for 3D human pose and shape modeling and extend it to modeling animals.

Specifically we learn a generative model of the 3D pose
and shape of animals and then fit this model to 2D image data as
illustrated in Fig.~\ref{fig:teaser}.
We focus on a subset of four-legged mammals that all have the same
number of ``parts'' and model members of  the families Felidae, Canidae, Equidae, Bovidae, and Hippopotamidae.
Our goal is to build a statistical shape model like SMPL
\cite{SMPL:2015}, which
captures human body shape variation in a low-dimensional Euclidean
subspace, models the articulated structure of the body, 
and can be fit to image data \cite{Bogo:ECCV:2016}.

Animals, however, differ from humans in several important ways.
First, the shape variation across species far exceeds the kinds of variations seen between humans.
Even within the canine family, there is a huge variability in dog shapes as a result of selective breeding.
Second, all these animals have tails, which are highly deformable and obviously not present in human shape models.
Third,  obtaining 3D data to train a model is much more challenging.
SMPL and previous models like it (e.g.~SCAPE \cite{Anguelov05}) rely on a large database of thousands of 3D scans of many people (capturing shape variation in the population) and a wide range of poses (capturing pose variation).
Humans are particularly easy and cooperative subjects.
It is impractical to bring a large number of wild animals into a lab environment for scanning and it would be difficult to take scanning equipment into the wild to capture animals shapes in nature.

\begin{figure*}[t]
\begin{center}
\includegraphics[width=1.0\linewidth]{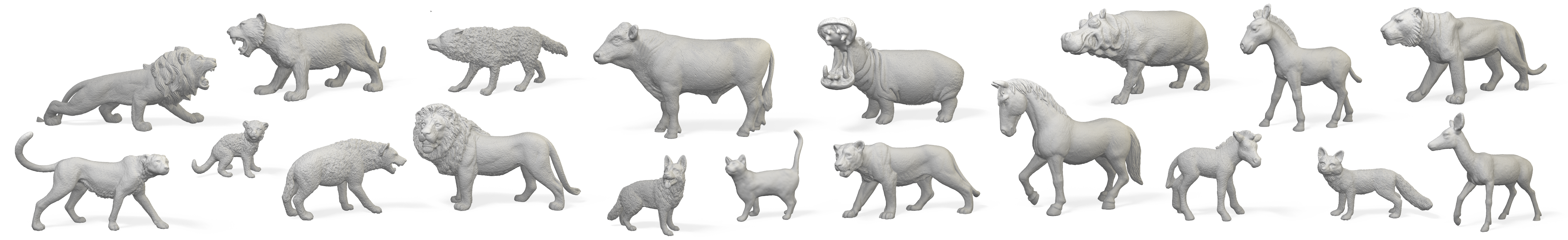}
\end{center}
\vspace{-1.5em}
\caption{{\bf Toys.}  Example 3D scans of animal figurines used for
  training our model.}
\vspace{-1.5em}
\label{fig:toys}
\end{figure*}

Since scanning live animals is impractical we instead scan
realistic toy animals to create a dataset of $41$ scans of a range of quadrupeds
as illustrated in Fig.~\ref{fig:toys}. 
We show that a model learned from toys generalizes to real animals.

The key to building a statistical 3D shape model is that all the 3D data must be in correspondence.
This involves registering a common template mesh to every scan.
This is a hard problem, which we approach by introducing a 
novel part-based model and
inference scheme that extends the ``stitched puppet'' (SP) model
\cite{Zuffi:CVPR:2015}. Our new {\em Global-Local Stitched Shape} model ({\em
  GLoSS}) 
aligns a template to different shapes, providing a coarse registration between very different
animals (Fig.~\ref{fig:regres} left).
The GLoSS registrations are somewhat crude but provide a reasonable initialization for a {\em model-free} refinement,
where the template mesh vertices deform towards the scan surface under an
As-Rigid-As-Possible (ARAP) constraint~\cite{Sorkine:EG:2007}
(Fig.~\ref{fig:regres} right).

Our template mesh is segmented into parts, with blend weights, so that
it can be reposed using linear blend skinning (LBS).
We ``pose normalize'' the refined registrations and learn a
low-dimensional shape space using principal component analysis (PCA).
This is analogous to the SMPL shape space \cite{SMPL:2015}. 

Using the articulated structure of the template and its blend weights, we obtain a model where new shapes can be generated and reposed. 
With the learned shape model, we refine the registration of the template to the scans using
co-registration \cite{Hirshberg:ECCV:2012}, which regularizes the
registration by penalizing deviations from the model fit to the scan.
We update the shape space and iterate to convergence.

The final {\em Skinned Multi-Animal Linear} model 
({\em SMAL}) provides a shape space of animals trained from 41 scans.
Because quadrupeds have shape variations in common,
the model generalizes to new animals not seen in training.
This allows us to fit SMAL to 2D data using manually detected keypoints
and segmentations. 
As shown in Fig.~\ref{fig:teaser} and Fig.~\ref{fig:Fits}, our model can generate realistic animal shapes
in a variety of poses.

In summary we describe a method to create a realistic 3D model of animals
and fit this model to 2D data.
The problem is much harder than modeling humans and we develop new
tools to extend previous methods to learn an animal model.
This opens up new directions for research on animal shape and motion capture.

\section{Related Work}
There is a long history on representing, classifying, and analyzing animal shapes in 2D \cite{Thompson}.
Here we focus only on work in 3D.
The idea of part-based 3D models of animals also has a long history.
Similar in spirit to our GLoSS model,
Marr and Nishihara \cite{Marr78} suggested that a wide range of animals shapes could be modeled by a small set of 3D shape primitives connected in a kinematic tree.  

{\bf Animal shape from 3D scans.}  
There is little work that systematically addresses the 3D scanning \cite{digitallife} and modeling of animals.  
The range of sizes and shapes, together with the difficulty of handling live animals and dealing with their movement, makes traditional scanning difficult.
Previous 3D shape datasets like TOSCA \cite{TOSCA} have a limited set of 3D animals that are artist-designed and with limited realism.

{\bf Animal shape from images.}
Previous work on modeling animal shape starts from the assumption that obtaining 3D animal scans is impractical
and focuses on using image data to extract 3D shape.
Cashman and Fitzgibbon \cite{dolphins} take a template of a dolphin 
and learn a low-dimensional model of its deformations from hand
clicked keypoints and manual segmentation. 
They optimize their model to minimize reprojection error to the keypoints and contour.
They also show results for a pigeon and a polar bear.
The formulation is elegant but the approach suffers from an overly smooth shape representation; this is not so problematic for dolphins but for other animals it is.
The key limitation, however, is that they do not model articulation. 

Kanazawa et al.~\cite{Kanazawa:Cats:2016} deform a 3D animal template 
to match hand clicked points in a set of images.
They learn separate deformable models for cats and horses using spatially varying stiffness values. 
Our model is stronger in that it captures articulation separately from shape variation.
Further we model the shape variation across a wide range of animals to produce a statistical shape model.

Ntouskos et al.~\cite{Ntouskos2015} take multiple views of different animals from the same class,
manually segment the parts in each view, and then fit geometric primitives to
segmented parts.
They assemble these to form an approximate 3D shape.
Vicente and Agapito \cite{Vicente2013} extract a template from a reference image and then deform it to fit a new image using keypoints and the silhouette.  
The results are of low resolution when applied to complex shapes.

Our work is complementary to these previous approaches that only use
image data to learn 3D shapes. Future work should combine 3D scans with image data to obtain even richer models. 

{\bf Animal shape from video.}
Ramanan et al.~\cite{Ramanan2006} model animals as a 2D kinematic chain of parts and learn the parts and their appearance from video.
Bregler et al.~\cite{Bregler2000} track features on a non-rigid object (e.g. a giraffe neck) and extract a 3D surface as well as its low-dimensional modes of deformation.
Del Pero et al.~\cite{DelPero2017} track and segment animals in video but do not address 3D shape reconstruction.
Favreau et al.~\cite{Favreau2004} focus on animating a 3D model of an animal given a 2D video sequence.  
Reinert et al.~\cite{Reinert:2016} take a video sequence of an animal
and, using an interactive sketching/tracking approach,  extract a textured 3D model of the
animal. The 3D shape is obtained by fitting generalized cylinders to each
sketched stroke over multiple frames. 

None of these methods model the kinds of detail available in 3D scans, nor do they model the 3D articulated structure of the body.
Most importantly none try to learn a 3D shape space spanning multiple animals.

{\bf Human shape from 3D scans.}
Our approach is inspired by a long history of learning 3D shape models of humans.
Blanz and Vetter \cite{blanz09} began this direction by aligning 3D scans of faces and computing a low-dimensional shape model.
Faces have less shape variability and are less articulated than animals, simplifying mesh registration and modeling.
Modeling articulated human body shape is significantly harder but several models have been proposed
\cite{Allen:2006:LCM,Anguelov05,tenbo,hasler2009statistical,SMPL:2015}.
Chen et al.~\cite{Chen2010} model both humans and sharks, factoring deformations into pose and shape.
The 3D shark model is learned from synthetic data and they do not model articulation.
Khamis et al.~\cite{khamis-cvpr2015} learn an articulated hand model with shape variation from depth images.

We base our method on SMPL \cite{SMPL:2015}, which combines a low-dimensional shape space with an articulated blend-skinned model.
SMPL is learned from 3D scans of 4000 people in a common pose and another 1800 scans of 60 people in a wide variety of poses.
In contrast, we have much less data and at the same time much more shape variability to represent.
Despite this, we show that we can learn a useful animal model for computer vision applications.
More importantly, this provides a path to making better models using more scans as well as image data.
We also go beyond SMPL to add a non-rigid tail and more parts than are present in human bodies.

\section{Dataset}
We created a dataset of 3D animals by scanning toy figurines
(Fig. \ref{fig:toys}) using an Artec hand-held 3D scanner. 
We also tried scanning taxidermy animals in a museum but found,
surprisingly, that the shapes of the toys looked more realistic. 
We collected a total of $41$ scans from several species: $1$ cat, $5$ cheetahs, $8$ lions, $7$ tigers, $2$ dogs, $1$ fox, $1$ wolf, $1$ hyena, $1$ deer, $1$ horse, $6$ zebras, $4$ cows, $3$ hippos.
We estimated a scaling factor so animals from different manufacturers were comparable in size.
Like previous 3D human datasets \cite{CAESAR}, and methods that create
animals from images \cite{dolphins,Kanazawa:Cats:2016}, we collected a
set of 36 hand-clicked keypoints that we use to aid mesh registration.
For more information see \cite{supmat}.

\section{Global/Local Stitched Shape Model}
\label{sec:gloss}

\begin{figure}[t]
\centerline{
   \includegraphics[width=0.7\linewidth]{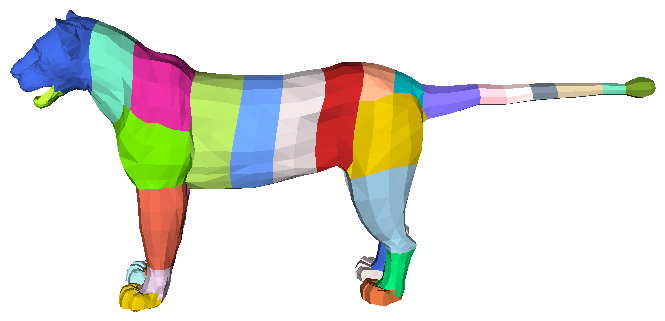}
}
   \caption{\textbf{Template mesh}. It is segmented into 33 parts, and here posed in the neutral 
     pose.}
\vspace{-1.5em}
\label{fig:segm}
\end{figure}

The Global/Local Stitched Shape model (GLoSS) is a 3D articulated model where body shape deformations are locally defined for each part
and the parts are assembled together by minimizing a stitching cost at the part interfaces. 
The model is inspired by the SP model \cite{Zuffi:CVPR:2015}, but has significant differences from it. 
In contrast to SP, the shape deformations of each part are analytic, rather than learned.
This makes it more approximate but, importantly, allows us to apply it to novel animal shapes, without requiring {\em a priori} training data.
Second, GLoSS is a globally differentiable model that can be fit to data with gradient-based techniques. 

To define a GLoSS model we need the following: 
a 3D template mesh of an animal with the desired polygon count, 
its segmentation into parts, skinning weights, and an animation sequence. 
To define the mesh topology, we use a 3D mesh of a lioness downloaded from the Turbosquid website.
The mesh is rigged and skinning weights are defined. 
We manually segment the mesh into $N=33$ parts (Fig.~\ref{fig:segm})
and make it symmetric along its sagittal plane. 
We now summarize the GLoSS  parametrization. 
Let $i$ be a part index, $i \in (1 \cdots\ N)$.
The model variables are: part location $\textbf{l}_i \in \mathbb{R}^{3\times1}$, part absolute 3D rotation $\textbf{r}_i \in \mathbb{R}^{3\times1}$, expressed as a Rodrigues vector, intrinsic shape variables $\textbf{s}_i \in \mathbb{R}^{n_{s}\times1}$ and pose deformation variables $\textbf{d}_i \in \mathbb{R}^{n_{d}\times1}$. 
Let $ \pi_i = \{ \textbf{l}_i, \textbf{r}_i, \textbf{s}_i, \textbf{d}_i \} $ be the set of variables for part $i$ and $\Pi = \{ \textbf{l}, \textbf{r}, \textbf{s}, \textbf{d} \} $ the set of variables for all parts.
The vector of vertex coordinates, $\hat{\textbf{p}}_i \in
\mathbb{R}^{3\times n_i}$, for part $i$ in a global reference frame is computed as:
%
\begin{equation}
\hat{\textbf{p}}_i(\pi_i) = R(\textbf{r}_i) \textbf{p}_i+\textbf{l}_i,
\end{equation} 
where $n_i$ is the number of vertices in the part,
and $R \in SO(3)$ is the rotation matrix obtained from $\textbf{r}_i$. 
The $\textbf{p}_i \in \mathbb{R}^{3\times n_i}$ are points in a local coordinate frame, computed as:
\begin{equation}
\text{vec}(\textbf{p}_i)=\textbf{t}_i+\textbf{m}_{p,i}+B_{s,i}\textbf{s}_i+B_{p,i}\textbf{d}_i.
\end{equation} 
Here $\textbf{t}_i \in \mathbb{R}^{3n_i\times1}$ is the part template, $\textbf{m}_{p,i} \in \mathbb{R}^{3n_i\times1}$ is the vector of average pose displacements; $B_{s,i} \in \mathbb{R}^{3n_i\times n_s}$ is a matrix with columns representing a basis of intrinsic shape displacements, and $B_{p,i} \in \mathbb{R}^{3n_i\times n_d}$ is the matrix of pose dependent deformations. These deformation matrices are defined below.

\noindent\textbf{Pose deformation space.} 
We compute the part-based pose deformation space from examples. 
For this we use an animation of the lioness template using linear blend skinning (LBS).
Each frame of the animation is a pose deformation sample. 
We perform PCA on the vertices of each part in a local coordinate frame, obtaining a vector of average pose deformations $\textbf{m}_{p,i}$ and the basis matrix $B_{p,i}$. 

\noindent\textbf{Shape deformation space.} We define a synthetic shape space for each body part. 
This space includes 7 deformations of the part template, namely scale, 
scale along $x$, scale along $y$, scale along $z$, and three stretch deformations that are defined as follows. Stretch for $x$ does not modify the $x$ coordinate of the template points, while it scales the $y$ and $z$ coordinates in proportion to the value of $x$. Similarly we define the stretch for $y$ and $z$.
This defines a simple analytic deformation space for each part. We model the distribution of the shape coefficients as a Gaussian distribution with zero mean and diagonal covariance, where we set the variance of each dimension arbitrarily.

\section{Initial Registration}
The initial registration of the template to the scans is performed in two steps.
First, we optimize the GLoSS model with a gradient-based method. 
This brings the model close to the scan. 
Then, we perform a model-free registration of the mesh vertices to the scan using As-Rigid-As-Possible (ARAP) regularization~\cite{Sorkine:EG:2007} to capture the fine details.

\noindent{\bf GLoSS-based registration.} 
To fit GLoSS to a scan, we minimize the following objective: 
\begin{eqnarray}
\lefteqn{E(\Pi) = E_{m}(\textbf{d}, \textbf{s})+E_{stitch}(\Pi) +} \nonumber \\
& &E_{curv}(\Pi)+E_{data}(\Pi)
+E_{pose}(\textbf{r}),
\label{eq:gloss}
\end{eqnarray}
where
\begin{equation*}
E_m(\textbf{d}, \textbf{s}) = k_{sm} E_{sm}(\textbf{s}) + k_s \sum_{i=1}^N E_{s}(\textbf{s}_i) + k_d \sum_{i=1}^N E_{d}(\textbf{d}_i)
\end{equation*}
is a model term, where $E_{s}$ is the squared Mahalanobis distance from the
synthetic shape distribution and $E_{d}$ is a squared $L2$ norm.
The term $E_{sm}$ represents the constraint that symmetric parts should have similar shape deformations. 
We impose similarity between left and right 
limbs, front and back paws, and sections of the torso. 
This last constraint favors sections of the torso to have similar length. 

The stitching term $E_{stitch}$ is the sum of squared distances of the corresponding points at the interfaces between parts (cf.~\cite{Zuffi:CVPR:2015}).
Let $C_{ij}$ be the set of vertex-vertex correspondences between
part~$i$ and part~$j$. Then $E_{stitch}(\Pi) =$
\begin{eqnarray}
k_{st} \sum_{(i,j) \in \mathcal{C}} \sum_{(k,l) \in C_{ij}} \Vert \hat{\textbf{p}}_{i,k}(\pi_i)- 
\hat{\textbf{p}}_{j,l} (\pi_j)\Vert^2,
\end{eqnarray}
where $\mathcal{C}$ is the set of part connections. 
Minimizing this term favors connected parts. 

The data term is defined as: $E_{data}(\Pi) =$
\begin{align}
\label{eq:data}
k_{kp} E_{kp}(\Pi) + 
k_{m2s} E_{m2s}(\Pi)+k_{s2m} E_{s2m}(\Pi),
\end{align}
where $E_{m2s}$ and $E_{s2m}$ are distances from the model to the scan and from the scan to the model, respectively:
\begin{equation}
E_{m2s}(\Pi) = \sum_{i=1}^N \sum_{k=1}^{n_i}  \rho(\min_{\textbf{s} \in \mathcal{S}} \Vert \hat{\textbf{p}}_{i,k}(\pi_i) - \textbf{s} \Vert^2),
\end{equation}
\begin{equation}
E_{s2m}(\Pi) = \sum_{l=1}^S \rho(\min_{\hat{\textbf{p}}} \Vert \hat{\textbf{p}}(\Pi) - \textbf{s}_l \Vert^2), 
\end{equation}
where $\mathcal{S}$ is the set of $S$ scan vertices and $\rho$ is the Geman-McClure robust error function \cite{geman}.
The term $E_{kp}(\Pi)$ is a term for
matching model keypoints with scan keypoints, 
 and is defined as the sum of squared distances between corresponding keypoints. This term is important to 
enable matching between extremely different animal shapes.

The curvature term favors parts that have a similar pairwise relationship as those in
the template; $E_{curv}(\Pi) = $
  \begin{eqnarray*}
     k_c \sum_{(i,j)\in \mathcal{C}} \sum_{(k,l) \in C_{ij}} \bigl\vert \Vert \hat{\textbf{n}}_{i,k}(\pi_i)- 
      \hat{\textbf{n}}_{j,l} (\pi_j)\Vert^2 - 
      \Vert \hat{\textbf{n}}^{(t)}_{i,k}- \hat{\textbf{n}}^{(t)}_{j,l} \Vert^2 \bigr\vert,
  \end{eqnarray*}
where $\hat{\textbf{n}}_i$ and $\hat{\textbf{n}}_j$ are vectors of
vertex normals on part~$i$ and part~$j$, respectively. 
Analogous quantities on the template are denoted with a superscript
$(t)$. 
Lastly, $E_{pose}$ is a pose prior on the tail parts learned from
animations of the tail. 
The values of the energy weights are manually defined and kept constant for all the toys.

\begin{figure}[t]
\centerline{
\includegraphics[width=1.0\columnwidth]{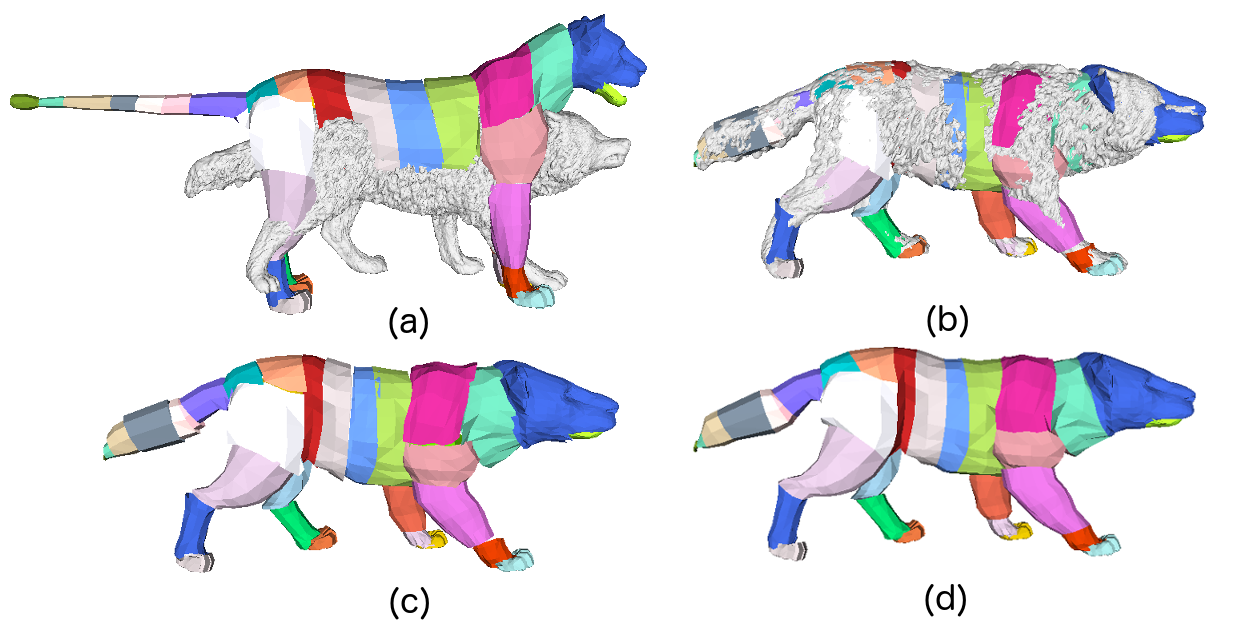}
}
\vspace{-.5em}
   \caption{\textbf{GLoSS fitting.} (a) Initial template and scan.  (b) GLoSS fit to
     scan.  (c) GLoSS model showing the parts.  (d) Merged mesh with global
     topology obtained by removing the duplicated vertices at the part interfaces.
}
\label{fig:optim}
\end{figure}

\begin{figure}[t]

  \begin{center}
   \includegraphics[width=\linewidth]{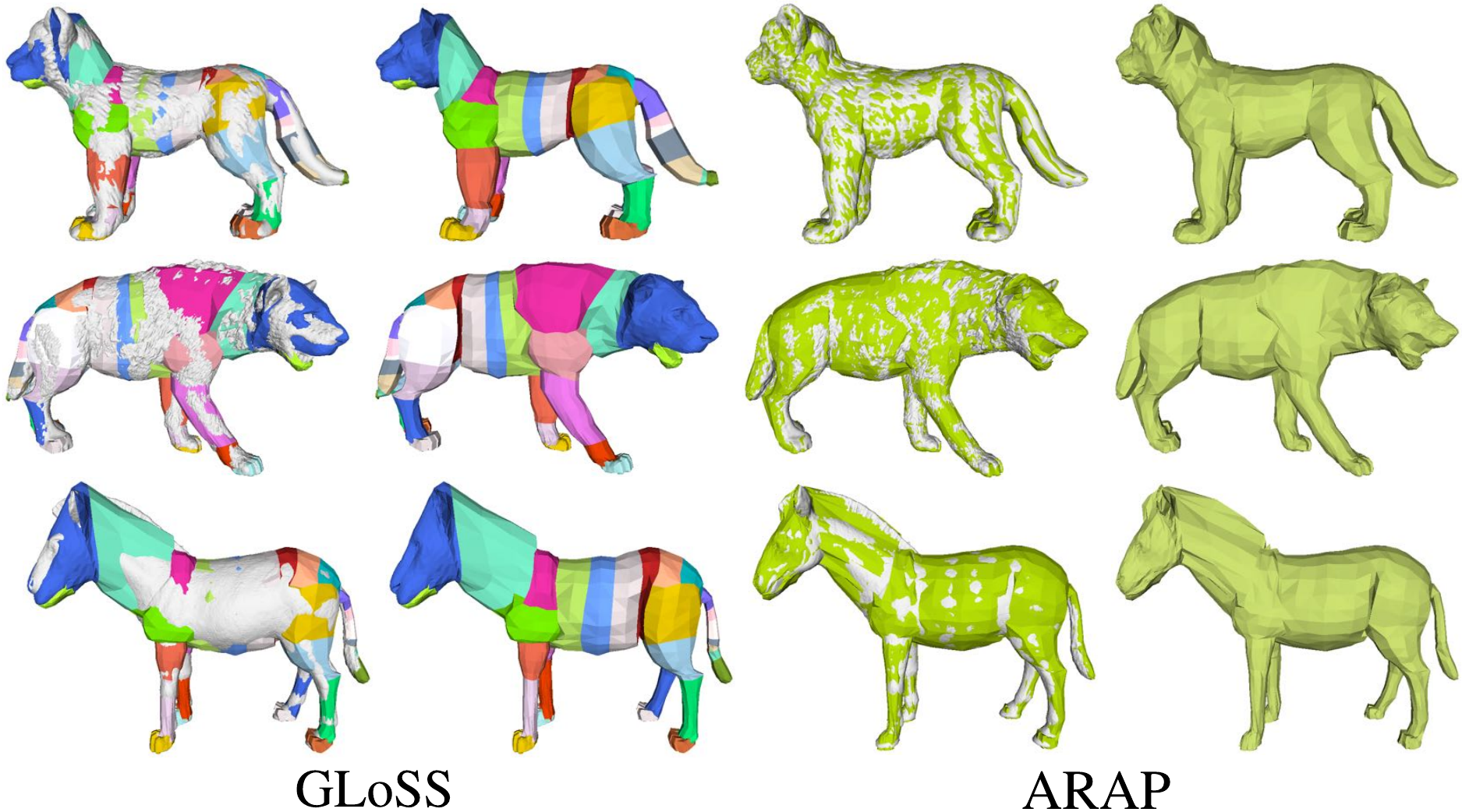}
  \end{center}
\vspace{-1.0em}
   \caption{\textbf{Registration results}. Comparing GLoSS (left) with the ARAP
     refinement (right).  The fit to the scan is much tighter after refinement.
}
\vspace{-.5em}
\label{fig:regres}
\end{figure}

We initialize the registration of each scan by aligning the model in neutral pose to the scan based on the median value of their vertices. 
Given this, we minimize Eq.~\ref{eq:gloss} using the Chumpy auto-differentiation package \cite{chumpy}.
Doing so aligns the lioness GLoSS model to all the toy scans.  
Figure~\ref{fig:optim}a-c shows an example of fitting of GLoSS (colored) to a scan (white), and Fig.~\ref{fig:regres} (first and second column) shows some of the obtained registrations. 
To compare the GLoSS-based registration with SP we computed SP registrations for the big cats family. We obtain an average scan-to-mesh distance of $4.39 (\sigma=1.66)$ for SP, and $3.22 (\sigma=1.34)$ for GLoSS.

\noindent{\bf ARAP-based refinement.} 
The GLoSS model gives a good initial registration.
Given this, we turn each GLoSS mesh from its part-based topology into a global
topology where interface points are not duplicated (Fig.~\ref{fig:optim}d). 
We then further align the vertices $\textbf{v}$ to the scans by
minimizing an energy function defined by a data term equal to Eq.~\ref{eq:data} and an As-Rigid-As-Possible (ARAP) regularization term~\cite{Sorkine:EG:2007}:
\begin{equation}
E(\textbf{v}) = E_{data}(\textbf{v})+E_{arap}(\textbf{v}).
\label{eq:earap}
\end{equation}
This model-free optimization brings the mesh vertices closer to the
scan and therefore  more accurately captures the shape of the animal
(see Fig.~\ref{fig:regres}).

\section{Skinned Multi-Animal Linear Model}

\begin{figure}[t]
\includegraphics[width=1.0\linewidth]{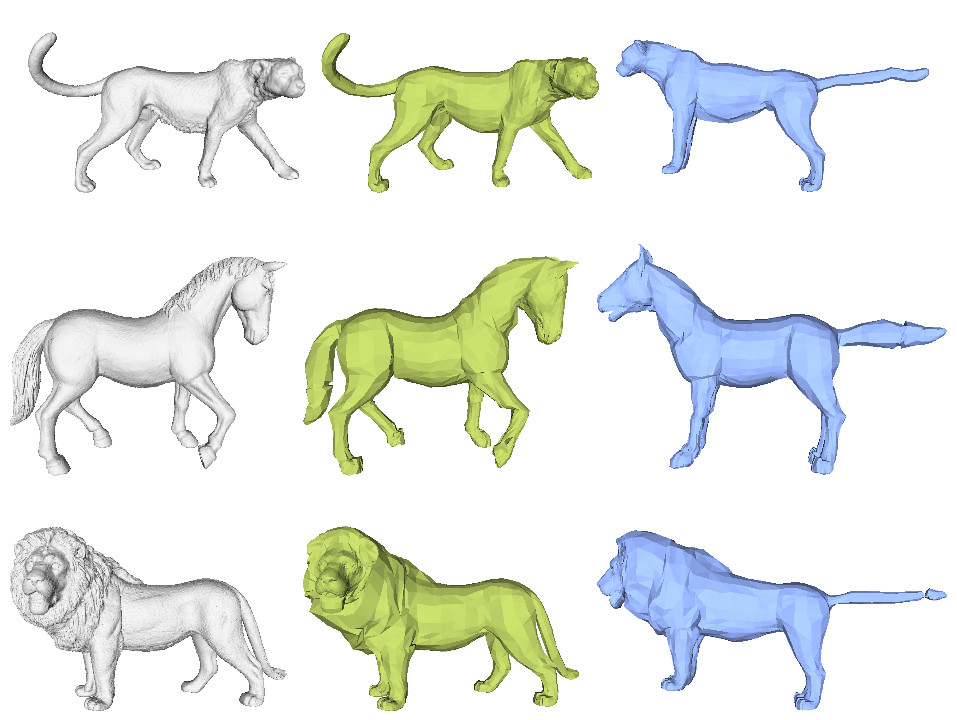}
\hspace*{0.4in} Scan \hspace*{0.5in}  ARAP \hspace*{0.5in}  Neutral pose
\vspace{-0.5em}
\caption{Registrations of toy scans in the neutral pose. }
\vspace{-1em}
\label{fig:tpose}
\end{figure}

The above registrations are now sufficiently accurate to create a first shape model, which we refine further below to produce the full SMAL model.

\noindent{\bf Pose normalization.}
Given the pose estimated with GLoSS, we bring all the registered
templates into the same neutral pose using LBS.
The resulting meshes are not symmetric. 
This is due to various reasons:
inaccurate pose estimation, limitations of linear-blend-skinning, the toys may not be symmetric, and pose differences across sides of the body create different deformations.
We do not want to learn this asymmetry.
To address this we perform an averaging of the vertices after we have
mirrored the mesh to obtain the registrations in the neutral pose (Fig.~\ref{fig:tpose}). 
Also, the fact that mouths are sometimes open and other times closed
presents a challenge for registration,
as inside mouth points are not observed in the scan when the animal has a closed mouth. To address this, palate and tongue points in the registration are regressed from the mouth points using a simple linear model learned from the template.
Finally we smooth the meshes with Laplacian smoothing. 

\noindent{\bf Shape model.}
Pose normalization removes the non-linear effects of part rotations on the vertices.  
In the neutral pose we can thus model the statistics of the shape variation in a Euclidean space.
We compute the mean shape and the principal components, which 
capture shape differences between the animals.

\noindent{\bf SMAL.} 
The SMAL model is a function $M(\myb, \myt, \myg)$ of shape
$\myb$, pose $\myt$ and translation $\myg$. $\myb$ is a vector of the coefficients
of the learned PCA shape space, $\myt \in \mathbb{R}^{3N} =
\{\textbf{r}_i\}_{i=1}^{N}$ is the relative rotation of the $N=33$ joints in the
kinematic tree, and $\myg$ is the global translation applied to the root
joint. 
Analogous to SMPL, the SMAL function returns a 3D mesh, where the template
model is shaped by $\myb$, articulated by $\myt$ through LBS, and shifted by
$\myg$. 

\noindent{\bf Fitting.}
To fit SMAL to scans we minimize the objective:
\begin{equation}
E(\myb, \myt, \myg) = E_{pose}(\myt)+E_{s}(\myb) + E_{data}(\myb, \myt, \myg),
\label{eq:smalfit}
\end{equation}
where $E_{pose}(\myt)$ and $E_{s}(\myb)$ are squared Mahalanobis distances from prior distributions for pose and shape, respectively. $E_{data}(\myb, \myt, \myg)$ is defined as in Eq.~\ref{eq:data} but over the SMAL model. 
For optimization we use Chumpy~\cite{chumpy}.

\noindent{\bf Co-registration.}
To refine the registrations and the SMAL model further, we then perform
co-registration \cite{Hirshberg:ECCV:2012}.
The key idea is to first perform a SMAL model optimization to align
the current model to the scans, and then run a model-free step where
we {\em couple}, or regularize, the model-free registration to the
current SMAL model by adding a coupling term to Eq.~\ref{eq:earap}:
\begin{equation}
E_{coup}(\textbf{v})=k_o \sum_{i=1}^{V} \vert \textbf{v}^0_i - \textbf{v}_i \vert, 
\end{equation}
where $V$ is the number of vertices in the template,
$\textbf{v}^0_i$ is vertex $i$ of the model fit to the scan, 
and the $\textbf{v}_i$ are the coupled mesh vertices being optimized.
During co-registration we use a shape space with $30$ dimensions. 
We perform $4$ iterations of registration and model building and
observe the registration errors decrease and converge
(see Sup.~Mat.~\cite{supmat}).
With the registrations to the toys in the last iteration we learn the shape space of our final SMAL model.

\begin{figure}[t]
\centerline{\includegraphics[width=1.0\linewidth]{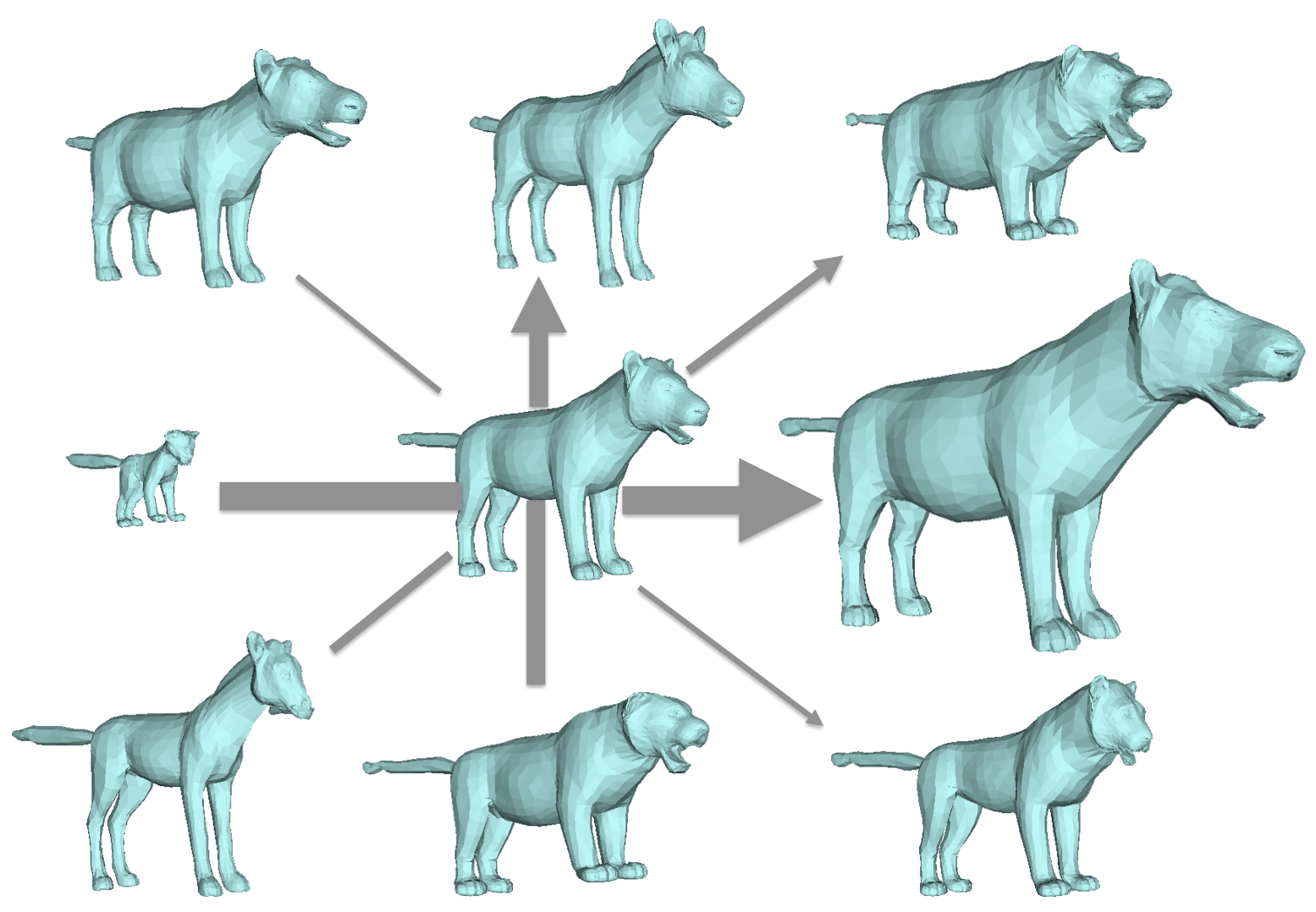} }
\vspace{-.5em}
\caption{\textbf{PCA space.} First 4 principal components.  Mean shape is in the
  center. The width of the arrow represents the order of the
  components. 
We visualise deviations of $\pm 2 \mathsf{std}$.}
\vspace{-.8em}
\label{fig:pcs}
\end{figure}

\begin{figure}[t]
\centerline{ \includegraphics[width=\linewidth]{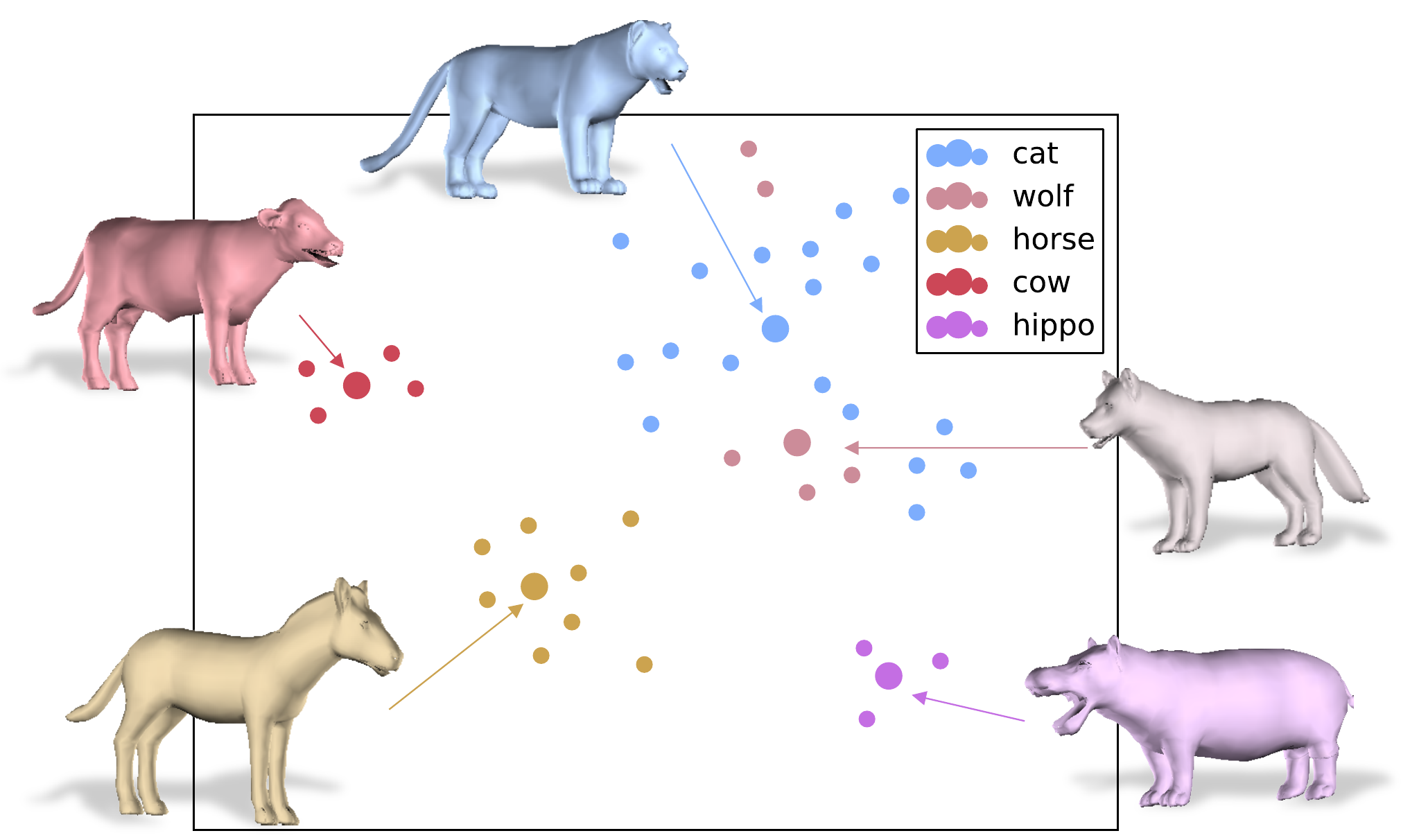}}
\vspace{-.5em}
   \caption{Visualization (using  t-SNE~\cite{tsne}) of different animal families using 8 PCs. Large dots indicate the mean of the PCA coefficients for each family.}
\vspace{-1em}
\label{fig:tsne}
\end{figure}

\noindent{\bf Animal shape space.}
After refining with co-registration, the final principal components are visualized in Fig.~\ref{fig:pcs}.
The global SMAL shape space captures the shape variability of
animals across different  families. 
The first component captures scale differences; our training set
includes adult and young animals. 
The learned space nicely separates shape characteristics of animal
families. This is illustrated in Fig.~\ref{fig:tsne} with a t-SNE visualization
\cite{tsne} of the first $8$ dimensions of the PCA coefficients in the training
set. The meshes correspond to the mean shape for each family. 
We also  define family-specific shape models by computing a Gaussian
over the PCA coefficients of the class.
We compare generic and family specific models below.

 \section{Fitting Animals to Images}
We now fit the SMAL model, $M(\myb, \myt, \myg)$, to image cues by optimizing the shape and pose parameters.
We fit the model to a combination of 2D keypoints and 2D silhouettes,
both manually extracted, as in previous work \cite{dolphins,Kanazawa:Cats:2016}.

We denote $\Pi(\cdot; f)$ as the perspective camera projection with focal
length $f$, where $\Pi(\textbf{v}_i; f)$ is the projection of the $i$'th vertex onto the
image plane and $\Pi(M; f) = \hat S$ is the projected model silhouette. We assume
an identity camera placed at the origin and that the global rotation of the 3D mesh
is defined by the rotation of the root joint.

To fit SMAL to an image, we formulate an objective function and
minimize it with respect to $\Theta=\{\myb, \myt, \myg, f\}$. 
The function is a sum of the keypoint and silhouette reprojection errors, a
shape prior, and two pose priors, $E(\Theta)=$\begin{equation}
E_{kp}(\Theta; \vx) + E_{silh}(\Theta; S) + E_{\myb}(\myb) + E_{\myt}(\myt) + E_{lim}(\myt).
  \label{eq:fit}
\end{equation} Each energy term is weighted by a hyper-parameter defining their
importance. 

\noindent{\bf Keypoint reprojection.} 
See \cite{supmat} for a definition of  keypoints which include surface
points and joints.
Since keypoints may be ambiguous, 
we assign a set of up to four vertices to represent each model keypoint and take the average
of their projection to match the target 2D keypoint. 
Specifically for the $k$'th keypoint, let $\vx$ be the labeled 2D keypoint
and $\{\textbf{v}_{k_j}\}_{j=1}^{k_m}$ be the assigned set of vertices, then
\begin{equation}
  \label{eq:kp}
E_{kp}(\Theta) = \sum_k \rho(||\vx - \frac{1}{|k_m|}\sum_{j=1}^{|k_m|}\Pi(\textbf{v}_{k_j}; \Theta)||_2),
\end{equation}
where $\rho$ is the Geman-McClure robust error function \cite{geman}.

\noindent{\bf Silhouette reprojection.} We encourage silhouette coverage
and consistency similar to \cite{Kar:CVPR:2015,Lassner:CVPR:2016,Weiss:ICCV:11} using a bi-directional distance:
 \begin{equation}
  \label{eq:silh}
  E_{silh}(\Theta) ={} \sum_{\vx \in \hat S}\mathcal{D_S}(\vx) +
  \sum_{\vx\in S} \rho(\min_{\hat{\vx} \in \hat S} ||\vx - \hat{\vx} ||_2),
\end{equation}
where $S$ is the ground truth silhouette and $\mathcal{D}_S$ is its L2 distance
transform field such that if point $\vx$ is inside the silhouette,
$\mathcal{D}_S(\vx)=0$.
Since the silhouette terms have small basins of attraction 
we optimize the term over multiple scales in a coarse-to-fine manner.

\noindent{\bf Shape prior.} 
We encourage $\myb$ to be close to the prior distribution of shape coefficients by defining $E_{\myb}$ to be the squared Mahalanobis distance with zero mean and
variance given by the PCA eigenvalues.
When the animal family is known, we can make our fits more specific by using the mean and covariance of training samples of the particular family.

\noindent{\bf Pose priors.} $E_{\myt}$ is also defined as the squared Mahalanobis
distance using the mean and covariance of the poses across all the training
samples and a walking sequence.
To make the pose prior symmetric,
we double the training data by reflecting the poses along the template's sagittal plane. Since
we do not have many examples, we further constrain the pose with limit bounds:
\begin{equation}
  \label{eq:lim}
  E_{lim}(\myt) = \max(\myt - \myt_{\max}, 0) + \max(\myt_{\min} - \myt_, 0).
\end{equation}
$\myt_{\max}$ and $\myt_{\min}$ are the maximum and minimum range of
values for each dimension of $\myt$ respectively, which we define by
hand. 
We do not limit the global rotation.

\noindent{\bf Optimization.}
Following \cite{Bogo:ECCV:2016}, we first initialize the depth of
$\myg$  using the torso points. 
Then we solve for the global rotation $\{\myt_i\}_{i=0}^3$ and $\myg$ using $E_{kp}$ over points on the torso. Using
these as the initialization, we solve Eq.~\ref{eq:fit} for the entire $\Theta$ without
$E_{silh}$. Similar to previous methods \cite{Bogo:ECCV:2016,
  Kanazawa:Cats:2016} we employ a staged approach where the weights on pose and shape
priors are gradually lowered over three stages. 
This helps avoid getting trapped in local optima.
We then finally include the $E_{silh}$ term 
and solve Eq.~\ref{eq:fit} starting from this initialization.
Solving for the focal length is important and
we regularize $f$ by adding another term that
forces $\myg$ to be close to its initial estimate. The entire optimization is done using OpenDR and Chumpy
\cite{chumpy,loper14}. 
Optimization for a single image typically takes less than a minute
on a common Linux machine. 

\begin{figure*}[t!]
\begin{center}
\includegraphics[width=1.0\linewidth]{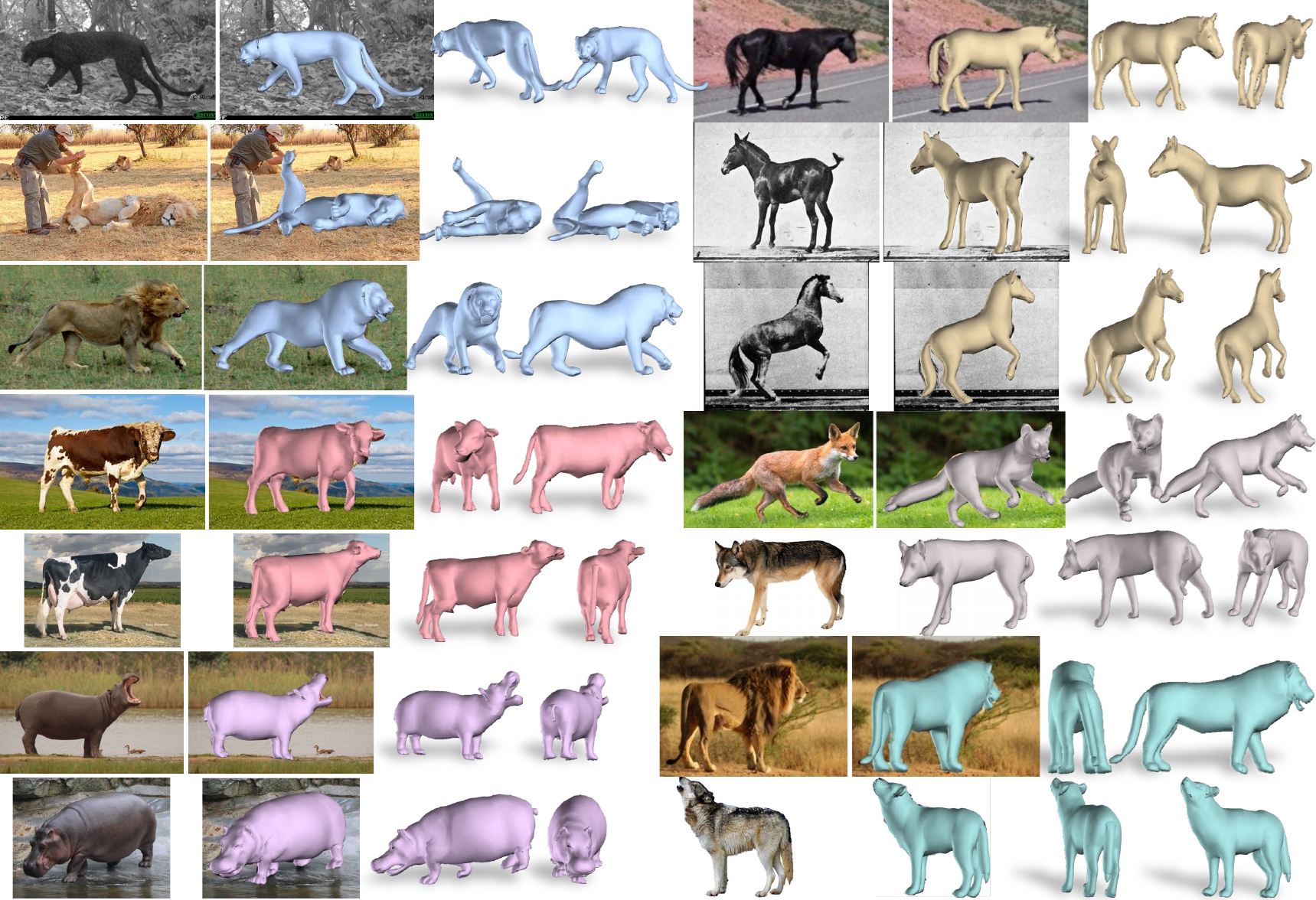}
\end{center}
\vspace{-1.5em}
   \caption{Fits to real images using manually obtained 2D points and
     segmentation. Colors indicate animal family. We show the input
     image, fit overlaid,
     views from $-45^{\circ}$ and  $45^{\circ}$. All results except
     for those in mint colors use the animal specific shape prior. The
     SMAL model, learned form toy figurines, generalizes to real
     animal shapes. }
\vspace{-.5em}
\label{fig:Fits}
\end{figure*}

\section{Experiments}
We have shown how to learn a SMAL animal model from a small set of toy figurines. Now the
question is: does this model capture the shape variation of real animals?
Here we test this by
fitting the model to annotated images of real animals. 
We fit using class specific and generic shape models, and show that
the shape space generalizes to new animal families not present in
training (within reason).

\noindent {\bf Data.}
For fitting, we use 19 semantic keypoints of \cite{DelPero2017} plus an
extra point for the tail tip. Note that these keypoints differ from those used in the 3D alignment. 
We fit frames in the TigDog dataset, reusing their
annotation, frames from the Muybridge footage, and images downloaded from the Internet. For images without annotation, we click the same 20 keypoints for
all animals, which takes about one minute for each image. 
We also hand segmented all the images.
No images were re-visited to improve their
annotations and we found the model to be robust to noise in the exact
location of the keypoints. 
See \cite{supmat} for data, annotations, and results.

\noindent{\bf Results.}
The model fits to real images of animals are shown in
Fig.~\ref{fig:teaser} and \ref{fig:Fits}. The weights for each term in
Eq.~\ref{eq:fit} are tuned by hand and held fixed for fitting \emph{all}
images. 
All results use the animal specific shape space except for those in mint
green, which use the generic shape model. 
Despite being trained on scans of toys, our model generalizes to
images of real animals, capturing their shape well. 
Variability in animal families with extreme shape characteristics (\eg lion manes, skinny horse legs, hippo faces) are modeled
well. 
Both the generic and class-specific models capture the shape of
  real animals well. 

\begin{figure}[t]
\begin{center}
\includegraphics[width=1.0\linewidth]{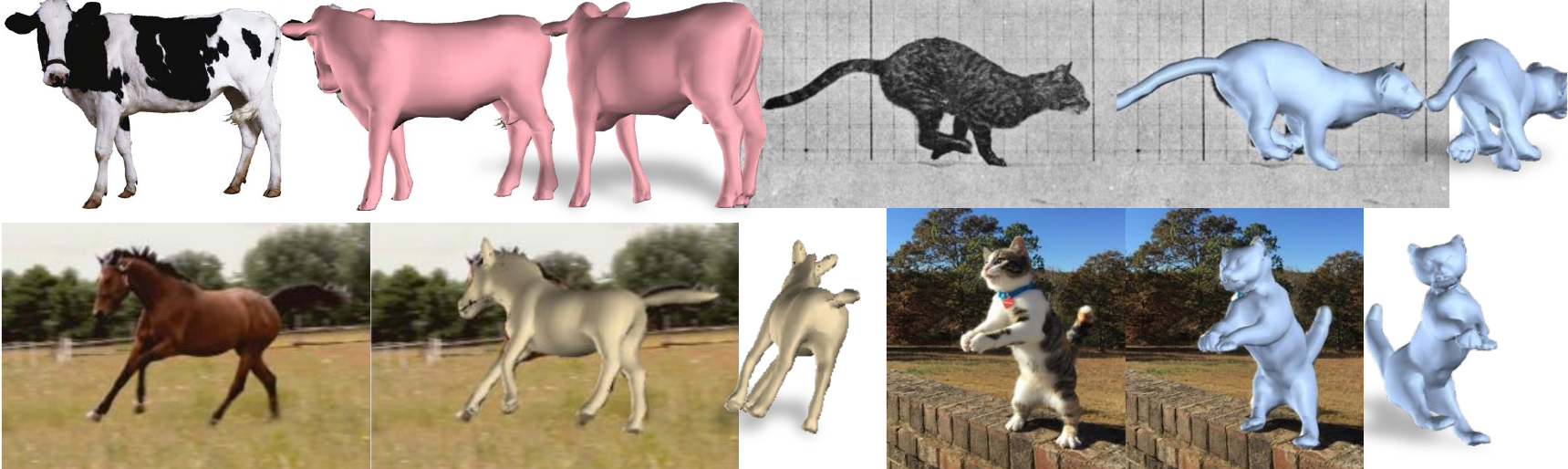}
\end{center}
\vspace{-1em}
   \caption{Failure examples due to depth ambiguity in pose and global rotation.}
\vspace{-1.5em}
\label{fig:Fail}
\end{figure}

\begin{figure}[t]
\begin{center}
\includegraphics[width=1.0\linewidth]{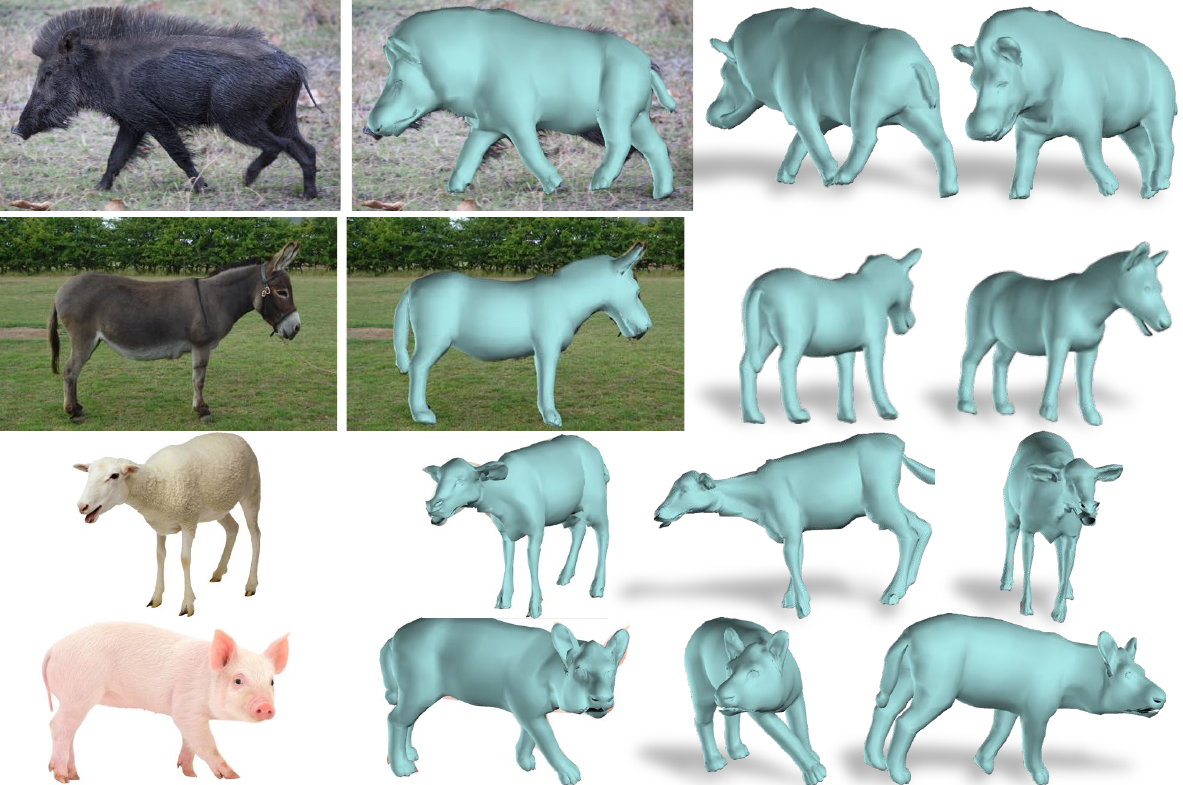}
\end{center}
   \caption{Generalization of SMAL to animal species not present in the
     training set.}
\label{fig:gen}
\end{figure}

Similar to the case of humans \cite{Bogo:ECCV:2016}, our main failures
are due to inherent depth ambiguity, both in global rotation and pose (Fig~\ref{fig:Fail}).
In  Fig.~\ref{fig:gen} we show the results of fitting the generic shape
model to classes of animals not seen in the training set: boar,
donkey, sheep and pigs.
While characteristic shape properties such as the pig snout cannot be exactly captured, these fits
suggest that the learned PCA space can generalize to new animals
within a range of quadrupeds.

\section{Conclusions}

Human shape modeling 
has a long history, while animal modeling is in its
infancy.
We have made small steps towards making the building of animal models
practical.
We showed that starting with toys, we can learn a model that
generalizes to images of real animals as well as to  
 types of animals not seen during training.
This gives a procedure for building richer models from more animals
and more scans.
While we have shown that toys are a good starting point, we would
clearly like a much richer model.
For that we believe that we need to incorporate image and video
evidence.
Our fits to images provide a starting point from which to learn 
richer deformations to explain 2D image evidence.
Here we have focused on a limited set of quadrupeds. 
A key issue is dealing with varying numbers of parts (e.g.~horns, tusks, trunks) and
parts of widely different shape (e.g.~elephant ears).
Moving beyond the class of animals here will involve creating a vocabulary of reusable shape parts and new ways of composing them.

\paragraph{Acknowledgements.}
We thank Seyhan Sitti for scanning the toys and Federica Bogo and Javier Romero for
their help with the silhouette term.  AK and DJ were supported by
the National Science Foundation under grant no. IIS-1526234.

{\small
\bibliographystyle{ieee}
\bibliography{refs}

\begin{thebibliography}{10}\itemsep=-1pt

\bibitem{supmat}
\url{http://smal.is.tue.mpg.de}.

\bibitem{chumpy}
\url{http://chumpy.org}.

\bibitem{digitallife}
Digital life.
\newblock \url{http://www.digitallife3d.com/}, Accessed November 12, 2016.

\bibitem{Allen:2006:LCM}
B.~Allen, B.~Curless, Z.~Popovi\'{c}, and A.~Hertzmann.
\newblock Learning a correlated model of identity and pose-dependent body shape
  variation for real-time synthesis.
\newblock In {\em Proceedings of the 2006 ACM SIGGRAPH/Eurographics Symposium
  on Computer Animation}, SCA '06, pages 147--156, Aire-la-Ville, Switzerland,
  Switzerland, 2006. Eurographics Association.

\bibitem{Anguelov05}
D.~Anguelov, P.~Srinivasan, D.~Koller, S.~Thrun, J.~Rodgers, and J.~Davis.
\newblock {SCAPE: Shape Completion and Animation of PEople}.
\newblock {\em ACM Trans.~Graph. (Proc.~SIGGRAPH}, 24(3):408--416, 2005.

\bibitem{blanz09}
V.~Blanz and T.~Vetter.
\newblock A morphable model for the synthesis of 3{D} faces.
\newblock In {\em SIGGRAPH}, pages 187--194. ACM, 1999.

\bibitem{Bogo:ECCV:2016}
F.~Bogo, A.~Kanazawa, C.~Lassner, P.~Gehler, J.~Romero, and M.~J. Black.
\newblock Keep it {SMPL}: Automatic estimation of {3D} human pose and shape
  from a single image.
\newblock In {\em European Conf. on Computer Vision (ECCV)}, Oct. 2016.

\bibitem{Bregler2000}
C.~Bregler, A.~Hertzmann, and H.~Biermann.
\newblock Recovering non-rigid 3d shape from image streams.
\newblock In {\em CVPR}, pages 2:690--696, 2000.

\bibitem{TOSCA}
A.~Bronstein, M.~Bronstein, and R.~Kimmel.
\newblock {\em Numerical Geometry of Non-Rigid Shapes}.
\newblock Springer Publishing Company, 2008.

\bibitem{dolphins}
T.~J. Cashman and A.~W. Fitzgibbon.
\newblock What shape are dolphins? building 3d morphable models from 2d images.
\newblock {\em IEEE Transactions on Pattern Analysis and Machine Intelligence},
  35(1):232--244, Jan 2013.

\bibitem{Chen2010}
Y.~Chen, T.~Kim, and R.~Cipolla.
\newblock Inferring {3D} shapes and deformations from single views.
\newblock In {\em iProc. European Conf. on Computer Vision, Part III}, page
  300–313, 2010.

\bibitem{tenbo}
Y.~Chen, Z.~Liu, and Z.~Zhang.
\newblock Tensor-based human body modeling.
\newblock In {\em IEEE Conf.~on Computer Vision and Pattern Recognition
  (CVPR)}, pages 105--112, June 2013.

\bibitem{DelPero2017}
L.~Del~Pero, S.~Ricco, R.~Sukthankar, and V.~Ferrari.
\newblock Behavior discovery and alignment of articulated object classes from
  unstructured video.
\newblock {\em International Journal of Computer Vision}, 121(2):303--325,
  2017.

\bibitem{Favreau2004}
L.~Favreau, L.~Reveret, C.~Depraz, and M.-P. Cani.
\newblock Animal gaits from video.
\newblock In {\em Proceedings of the 2004 ACM SIGGRAPH/Eurographics symposium
  on Computer animation}, pages 277--286. Eurographics Association, 2004.

\bibitem{geman}
S.~Geman and D.~McClure.
\newblock Statistical methods for tomographic image reconstruction.
\newblock {\em Bulletin of the International Statistical Institute},
  52(4):5--21, 1987.

\bibitem{hasler2009statistical}
N.~Hasler, C.~Stoll, M.~Sunkel, B.~Rosenhahn, and H.~Seidel.
\newblock A statistical model of human pose and body shape.
\newblock {\em Computer Graphics Forum}, 28(2):337--346, 2009.

\bibitem{Hirshberg:ECCV:2012}
D.~Hirshberg, M.~Loper, E.~Rachlin, and M.~Black.
\newblock Coregistration: Simultaneous alignment and modeling of articulated
  {3D} shape.
\newblock In {\em European Conf. on Computer Vision (ECCV)}, LNCS 7577, Part
  IV, pages 242--255. Springer-Verlag, Oct. 2012.

\bibitem{Kanazawa:Cats:2016}
A.~Kanazawa, S.~Kovalsky, R.~Basri, and D.~Jacobs.
\newblock Learning 3d deformation of animals from 2d images.
\newblock {\em Comput. Graph. Forum}, 35(2):365--374, May 2016.

\bibitem{Kar:CVPR:2015}
A.~Kar, S.~Tulsiani, J.~Carreira, and J.~Malik.
\newblock Category-specific object reconstruction from a single image.
\newblock In {\em CVPR}, 2015.

\bibitem{khamis-cvpr2015}
S.~Khamis, J.~Taylor, J.~Shotton, C.~Keskin, S.~Izadi, and A.~Fitzgibbon.
\newblock Learning an efficient model of hand shape variation from depth
  images.
\newblock In {\em IEEE Conference on Computer Vision and Pattern Recognition},
  2015.

\bibitem{Lassner:CVPR:2016}
C.~Lassner, J.~Romero, M.~Kiefel, F.~Bogo, M.~J. Black, and P.~V. Gehler.
\newblock Unite the people: Closing the loop between {3D} and {2D} human
  representations.
\newblock In {\em Proc. of the {IEEE} Conference on Computer Vision and Pattern
  Recognition {CVPR}}, 2017.

\bibitem{loper14}
M.~Loper and M.~J. Black.
\newblock Open{DR}: An approximate differentiable renderer.
\newblock In {\em European Conf. on Computer Vision (ECCV)}, pages 154--169,
  2014.

\bibitem{SMPL:2015}
M.~Loper, N.~Mahmood, J.~Romero, G.~Pons-Moll, and M.~J. Black.
\newblock {SMPL}: A skinned multi-person linear model.
\newblock {\em ACM Trans. Graphics (Proc. SIGGRAPH Asia)}, 34(6):248:1--248:16,
  Oct. 2015.

\bibitem{tsne}
L.~v.~d. Maaten and G.~Hinton.
\newblock Visualizing data using t-sne.
\newblock {\em Journal of Machine Learning Research}, 9(Nov):2579--2605, 2008.

\bibitem{Marr78}
D.~Marr and K.~Nishihara.
\newblock Representation and recognition of the spatial organization of three
  dimensional shapes.
\newblock {\em Proceedings of the Royal Society of London. Series B, Biological
  Sciences}, 200(1140):269--294, 1978.

\bibitem{Ntouskos2015}
V.~Ntouskos, M.~Sanzari, B.~Cafaro, F.~Nardi, F.~Natola, F.~Pirri, and M.~Ruiz.
\newblock Component-wise modeling of articulated objects.
\newblock In {\em The IEEE International Conference on Computer Vision (ICCV)},
  December 2015.

\bibitem{Ramanan2006}
D.~Ramanan, D.~A. Forsyth, and K.~Barnard.
\newblock Building models of animals from video.
\newblock {\em Pattern Analysis and Machine Intelligence, IEEE Transactions
  on}, 28(8):1319--1334, 2006.

\bibitem{Reinert:2016}
B.~Reinert, T.~Ritschel, and H.-P. Seidel.
\newblock Animated 3d creatures from single-view video by skeletal sketching.
\newblock In {\em GI '16: Proceedings of the 42st Graphics Interface
  Conference}, 2016.

\bibitem{CAESAR}
K.~Robinette, S.~Blackwell, H.~Daanen, M.~Boehmer, S.~Fleming, T.~Brill,
  D.~Hoeferlin, and D.~Burnsides.
\newblock {Civilian American and European Surface Anthropometry Resource
  (CAESAR)} final report.
\newblock Technical Report AFRL-HE-WP-TR-2002-0169, {US Air Force Research
  Laboratory}, 2002.

\bibitem{Sorkine:EG:2007}
O.~Sorkine and M.~Alexa.
\newblock As-rigid-as-possible surface modeling.
\newblock In {\em Proceedings of the Fifth Eurographics Symposium on Geometry
  Processing, Barcelona, Spain, July 4-6, 2007}, pages 109--116, 2007.

\bibitem{Thompson}
D.~W. Thompson.
\newblock {\em On Growth and Form}.
\newblock Cambridge University Press, 1917.

\bibitem{Vicente2013}
S.~Vicente and L.~Agapito.
\newblock Balloon shapes: Reconstructing and deforming objects with volume from
  images.
\newblock In {\em Conference on 3D Vision-3DV}, 2013.

\bibitem{Weiss:ICCV:11}
A.~Weiss, D.~Hirshberg, and M.~Black.
\newblock Home {3D} body scans from noisy image and range data.
\newblock In {\em Int. Conf. on Computer Vision (ICCV)}, Barcelona, Nov. 2011.
  IEEE.

\bibitem{Zuffi:CVPR:2015}
S.~Zuffi and M.~J. Black.
\newblock The stitched puppet: A graphical model of {3D} human shape and pose.
\newblock In {\em IEEE Conf. on Computer Vision and Pattern Recognition (CVPR
  2015)}, pages 3537--3546, June 2015.

\end{thebibliography}
}

\end{document}